\documentclass[letterpaper, 10 pt, conference]{ieeeconf}  % Comment this line out if you need a4paper

\IEEEoverridecommandlockouts                              % This command is only needed if 
\usepackage{graphics} % for pdf, bitmapped graphics files
\usepackage{epsfig} % for postscript graphics files
\usepackage{mathptmx} % assumes new font selection scheme installed
\usepackage{times} % assumes new font selection scheme installed
\usepackage{amsmath} % assumes amsmath package installed
\usepackage{amssymb}  % assumes amsmath package installed

\usepackage[hidelinks]{hyperref}
\usepackage{subcaption}

\usepackage{nicefrac} % correct typesetting of in-text fractions
\usepackage{siunitx} % correct typesetting of SI units
\usepackage{booktabs} % tables that are APA conformant and readable
\usepackage[dvipsnames]{xcolor} % get some color if you need it
\usepackage{svg}

\title{
\small \vspace{-3em}
Cite as: F. M. López, M. R. Ernst, F. Cruz, M. Hoffmann, and J. Triesch, ``Infant Spontaneous Movement Noise Improves Exploration in Deep RL'', in \textit{2026 IEEE International Conference on Development and Learning (ICDL)}. IEEE, 2026, pp. 1–6. \\[1em]
\LARGE \bf
Infant Spontaneous Movement Noise Improves Exploration in Deep RL
}

\author{
    Francisco~M.~López$^{1}$, 
    Markus~R.~Ernst$^{2}$,
    Francisco~Cruz$^{3}$,
    Matej~Hoffmann$^{4}$,
    and~Jochen~Triesch$^{5}$%
    \thanks{This work was supported by the Deutsche Forschungsgemeinschaft (German Research Foundation, DFG) under Germany’s Excellence Strategy (EXC 3066/1 “The Adaptive Mind”, Project No. 533717223). M.H. was supported by the Czech Science Foundation (GA ČR), project no. 25-18113S. J.T. was supported by the Johanna Quandt foundation. M.R.E was supported by an University International Postgraduate Award (UIPA) provided by UNSW Sydney. We thank Jason Khoury and Filipe Gama for assistance with the preparation and processing of the infant videos. Experiments were performed on the UNSW Compute Cluster Katana (DOI:~\href{https://doi.org/10.26190/669X-A286}{10.26190/669X-A286}).}
	\thanks{$^{1}$Francisco M. López is with the Frankfurt Institute for Advanced Studies, Germany, and with the School of Computer Science and Engineering, University of New South Wales, Australia. {\tt\footnotesize lopez@fias.uni-frankfurt.de}}%
    \thanks{$^{2}$Markus R. Ernst is with the School of Computer Science and Engineering, University of New South Wales, Australia. {\tt\footnotesize m.ernst@unsw.edu.au}}%
    \thanks{$^{3}$Francisco Cruz is with the School of Computer Science and Engineering, University of New South Wales, Australia, and with Escuela de Ingenier\'ia, Universidad Central de Chile, Chile. {\tt\footnotesize f.cruz@unsw.edu.au}}%
    \thanks{$^{4}$Matej Hoffmann is with the Department of Cybernetics, Faculty of Electrical Engineering, Czech Technical University in Prague, Czech Republic. {\tt\footnotesize matej.hoffmann@fel.cvut.cz}}%
    \thanks{$^{5}$Jochen Triesch is with the Frankfurt Institute for Advanced Studies, Germany. {\tt\footnotesize triesch@fias.uni-frankfurt.de}}%
}

\begin{document}

\maketitle
\thispagestyle{empty}
\pagestyle{empty}

%%%%%%%%%%%%%%%%%%%%%%%%%%%%%%%%%%%%%%%%%%%%%%%%%%%%%%%%%%%%%%%%%%%%%%%%%%%%%%%%
\begin{abstract}

Exploration in deep reinforcement learning (RL) is commonly implemented as temporally uncorrelated white noise. However, recent works show that temporally correlated colored noise can improve exploration efficiency by producing smooth trajectories with better coverage of the state space. We inquire whether action noise inspired by infant spontaneous movements can also improve exploration in deep RL. We find that the power spectral densities of babies' end-effector velocities follow a colored noise process where the spectral exponent increases with age. Inspired by this developmental pattern, we introduce a mechanism that progressively increases the temporal auto-correlation of exploration noise during RL training, matching the infant statistics. Experiments across several RL environments show that infant-inspired noise produces structured exploratory behavior and can improve learning efficiency compared to conventional exploration strategies. These findings suggest that human motor and cognitive development can provide useful guidance for designing learning mechanisms in artificial agents. Our code is available at \href{https://github.com/trieschlab/baby-noise-rl}{https://github.com/trieschlab/baby-noise-rl}.

\end{abstract}

%%%%%%%%%%%%%%%%%%%%%%%%%%%%%%%%%%%%%%%%%%%%%%%%%%%%%%%%%%%%%%%%%%%%%%%%%%%%%%%%
\section{Introduction}

 Humans have a protracted postnatal helplessness period~\cite{cusack2024helpless} and have to learn many behaviors through trial-and-error interactions with their environments, typically with limited external supervision. Within the first year of life, an infant will typically learn to combine over 600 muscles in her body to stand up, walk, and grasp objects. This is no simple feat. An agent with just 50 degrees of freedom of motion, each with 10 discrete possible rotation angles, would have access to \(10^{50}\) unique body configurations. This vastly exceeds the number of milliseconds that have elapsed since the start of the Universe, estimated at \(5 \times 10^{20}\), let alone a single year. Therefore, babies must be using efficient strategies that allow them to explore and discover useful behaviors in little time. We propose that deep RL can benefit greatly by drawing inspiration from infants. 

In this work, we focus on one particular component of exploration in deep RL: action noise. It is common practice to perturb a policy with stochastic noise during training to promote exploration~\cite{lillicrap2016continuous,fujimoto2018addressing,haarnoja2018soft}. The default choice for most RL algorithms is temporally uncorrelated Gaussian (white) noise. Some studies have shown an advantage when using the strongly correlated Ornstein-Uhlenbeck (OU) noise~\cite{lillicrap2016continuous}. However, several recent works have shown that mildly correlated noise, also known as colored noise~\cite{haunggi1994colored}, can improve exploration efficiency by producing smooth exploration trajectories with better coverage of the state space~\cite{pinneri2021sample,hollenstein2022action,eberhard2023pink,hollenstein2024colored,hollenstein2024pink} (see Fig.~\ref{fig:showcase}). Colored noise can be generated by enforcing a power-law $S(f)\propto f^{-\beta}$ in the power spectral density (PSD) of the noise. The exponent \(\beta\) defines the noise color, i.e., the degree of temporal correlation. Some values are common throughout multiple disciplines and have names, e.g., \(\beta=0\) is white noise, \(\beta=1\) is pink noise, and \(\beta=2\) is red noise (also called Brownian noise). Pink noise, in particular, has been identified as a better default than white noise for exploration across a variety of RL algorithms and environments~\cite{eberhard2023pink,hollenstein2024pink}.

On the other hand, many biological signals display dynamics with PSDs of the form $S(f)\propto f^{-\beta}$. Human motor behaviors~\cite{newell1985coordination,raffalt2023stride,gilfriche2018frequency} as well as neural activities~\cite{he2014scale} exhibit long-range temporal auto-correlations. In the context of development, spontaneous infant movements have also been shown to exhibit complex temporal organization~\cite{thelen1994dynamic,smith2003development,hadders2018early}. These findings suggest that variability in biological systems is not unstructured but rather temporally organized.

\begin{figure*}[!t]
\centering
%\includesvg{figures/overview_alt2.svg}
\includegraphics[width=1.0\linewidth]{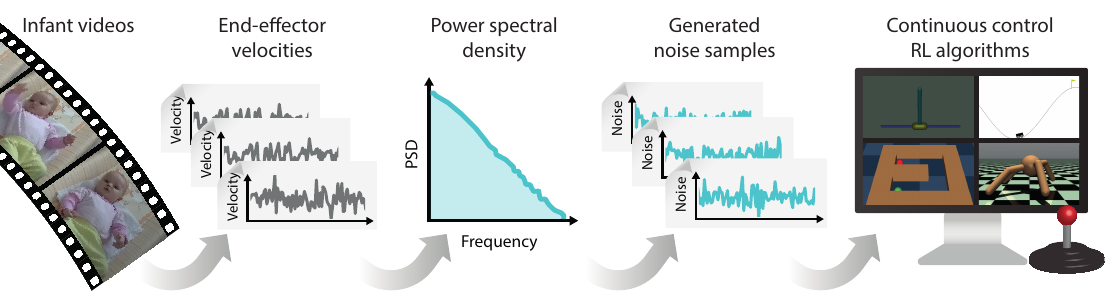}\\
\caption{Overview of our work. Spontaneous movement velocities are extracted from videos of infants and analyzed to obtain a developmental schedule of noise color that can be used to promote exploration in RL environments.}
\label{fig:overview}
\end{figure*}

In light of these findings, we investigate whether action noise inspired by infant spontaneous movements can also improve exploration in deep RL (see Fig.~\ref{fig:overview}). Our contributions are as follows: 
\begin{itemize}
    \item We quantify spontaneous movement noise as the PSD slope of end-effector velocities using 2D keypoints extracted from infant video recordings.
    \item We identify a developmental pattern where the noise color \(\beta\) increases with age between 8 and 30 weeks.
    \item We translate our findings into a developmentally scheduled colored action noise suitable for deep RL.
    \item We demonstrate that this infant-inspired exploration strategy improves learning performance across several control tasks. 
\end{itemize}

Our findings add to a growing body of work that suggests that motor and cognitive development of infants can provide useful guidance for designing efficient learning mechanisms in artificial agents~\cite{oudeyer2007intrinsic,zaadnoordijk2022lessons,aubret2023time}.

\section{Characterizing Infant Movement Noise} \label{sec:infant}

To characterize the temporal structure of infant spontaneous movements, we analyze a longitudinal dataset of video recordings collected during 19 sessions with four infants between the ages of 8 and 30 weeks. The infants were always recorded in the supine position while moving freely. The videos range in duration from 50 seconds to over 20 minutes. We use the OpenPose~\cite{cao2019openpose} human pose estimation method to extract the 2D positions of the end-effectors (wrists and ankles) within each video frame. Velocities are computed by taking temporal differences of the 2D keypoint coordinates, yielding a total of four time series, one per end effector. We then detrend and z-score each time series separately.

\begin{figure}[!t]
  \centering
    \begin{subfigure}{0.235\textwidth}
        \centering\includegraphics[width=\textwidth]{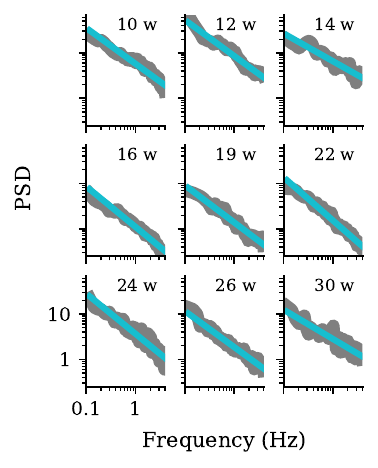}
        \caption{PSDs from a single infant.}
    \end{subfigure}
    \hfill
    \begin{subfigure}{0.235\textwidth}
        \centering\includegraphics[width=\textwidth]{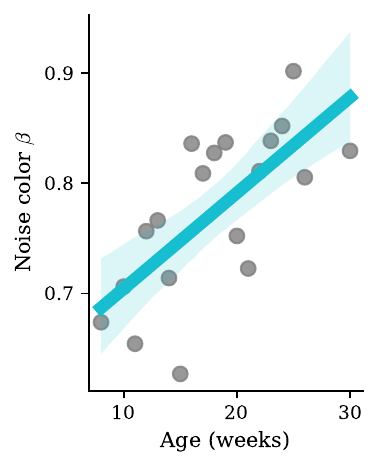}
        \caption{Development of noise color.}
    \end{subfigure}
    \caption{Behavioral results. A: Individual PSDs obtained from videos of a single infant between 10 and 30 weeks. The negative slopes from the linear regressions indicate the noise color \(\beta\) at each age. B: Development of noise color with age (\(R=0.70, p<0.001\)) from four infants across 19 sessions. This correlation reveals a gradual increase in the temporal auto-correlation of infant spontaneous movements.}
    \label{fig:behavior}
\end{figure}

To quantify the noise color, we perform a PSD analysis using Welch's method. Each time series is split into windows of 1{,}024 video frames (34.1 seconds with 30 Hz frame rate) and transformed into the frequency domain using a Fast Fourier Transform (FFT). We then take the average over all windows across all end-effectors to obtain a single PSD for each video recording. First, we verify that all PSDs obey a power-law relation $S(f)\propto f^{-\beta}$, where $S(f)$ denotes the power spectral density at frequency $f$ and $\beta$ is the spectral exponent characterizing the noise color. To estimate $\beta$, we perform a linear regression to the PSD in log-log space:
\begin{equation}
    \log(S) = -\beta \cdot \log(f) + c,
\end{equation}

\noindent where the negative slope returns a single noise exponent \(\beta\) for each session. The linear fits are computed over a frequency interval between \qty{0.1}{Hz} and \qty{4}{Hz} to avoid boundary effects and to minimize the influence of measurement noise introduced by OpenPose during keypoint extraction. For all 19 sessions we find strong correlations (\(p<0.001\)). Examples from one of the infants are shown in Fig.~\ref{fig:behavior}.A.

Finally, we analyze the evolution of the noise color during development. To do so, we perform a linear regression of the estimated $\beta$ values as a function of infant age in weeks (\(w\)):
\begin{equation} \label{eq:beta-infant}
    \beta(w) = a + b \cdot w.
\end{equation}

The results are shown in Fig.~\ref{fig:behavior}. We find a strong linear correlation (\(R=0.70, p<0.001\)) with intercept \(a=6.17\times10^{-1}\) and slope \(b=8.69\times10^{-3}\ \mathrm{weeks}^{-1}\), yielding a noise color \(\beta = 0.686\) at 8 weeks and \(\beta = 0.877\) at 30 weeks. To further validate this analysis, we repeat the entire procedure separately for the 4 end-effectors. In all cases we find significant correlations (\(p<0.01\)) between noise color and age with similar \(\beta\) values.

These results indicate that infant motor variability transitions toward more temporally structured dynamics during the first months of life. This finding is consistent with previous studies in developmental science suggesting that motor control evolves from relatively uncoordinated patterns toward organized and structured behaviors~\cite{thelen1994dynamic,smith2003development,hadders2018early}. However, the expression of motor development in terms of a quantitative change in noise color is an innovation of our work and allows for a straightforward application to exploration in artificial systems, including RL agents.

\section{Generating Infant-Inspired Action Noise} \label{sec:action-noise}

Deep RL algorithms require some form of exploration to learn successful behaviors. In continuous control, this can be achieved by means of an explicit action noise. Deterministic algorithms, including DDPG~\cite{lillicrap2016continuous} and its successor TD3~\cite{fujimoto2018addressing}, add random noise to the deterministic policy. For stochastic algorithms such as SAC~\cite{haarnoja2018soft}, the policy is sampled from a distribution which includes the action noise. While these algorithms typically rely on white noise, recent studies suggest that colored noise~\cite{pinneri2021sample,hollenstein2024colored}, and in particular pink noise with \(\beta=1\)~\cite{eberhard2023pink,hollenstein2024pink}, can improve learning performance. These works, however, make use of a constant noise color during training. Given the results shown in Section \ref{sec:infant}, we propose that a developmentally scheduled action noise with an annealing of the \(\beta\) parameter can likewise yield more efficient exploration strategies. Our proposal is detailed in the remainder of this section.

\subsection{Colored Noise Blocks}

Noise can be characterized using PSDs with a power law \(S(f) \propto f^{-\beta}\), with \(\beta=0\) for white noise and \(\beta>0\) for colored noise. Larger values of \(\beta\) produce smoother signals with stronger long-range temporal correlations. Because of these correlations, generating individual random samples with colored noise at each timestep can be too computationally expensive. Instead, long sequences of random colored noise samples can be generated in blocks using spectral shaping. We adapt the method presented by~\cite{eberhard2023pink}, itself adapted from~\cite{timmer1995generating}, as described next.

\begin{figure*}[!t]
\centering
\includegraphics[width=1.0\linewidth]{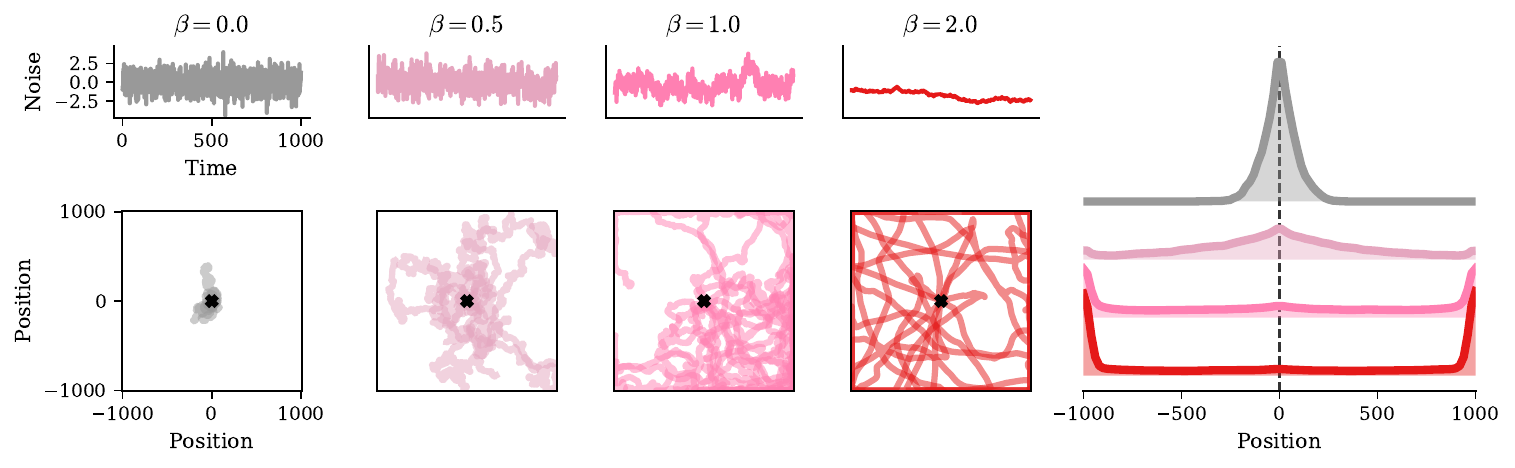}
\caption{Dynamics of different noise colors. Top: Examples of short time series generated through spectral shaping with noise colors \(\beta=0.0\) (white), \(\beta=0.5\) (blush), \(\beta=1.0\) (pink), and \(\beta=2.0\) (red). Bottom: Examples of \(10\) trajectories integrated from noise blocks of \(10{,}000\) samples in a 2D space of size \(2{,}000\). Right: Distributions of horizontal positions obtained from \(1{,}000\) trajectories on the same 2D space.}
\label{fig:showcase}
\end{figure*}

To generate a colored noise block, we first generate a sequence of Gaussian white noise samples and transform it to the frequency domain via the FFT. The frequency spectrum is then scaled to match the expected PSD and the resulting signal is transformed back into the time domain with the inverse FFT. This scaled signal preserves the variance of the original signal but has the temporal correlations corresponding to its noise color. The resulting block can be directly sampled to obtain colored action noise for a RL algorithm. Example sequences of colored noise are shown in Fig.~\ref{fig:showcase}, where, as expected, higher values of \(\beta\) are found to produce more temporally correlated signals.

We use a default length size of \(L = 10{,}000\) samples per block, which is long enough to guarantee a smooth PSD but also short enough that generating the colored noise does not result in any expensive computational overhead. As soon as a noise sequence is exhausted, a new one must be generated. For environments with more-than-one dimensional action spaces, an independent colored noise block is generated per action dimension.

As a proof-of-concept for the use of temporally correlated noise for exploration, we generate random 2D trajectories from colored noise. Fig.~\ref{fig:showcase} shows 10 such trajectories, each created by integrating a noise block of length \(10{,}000\) per dimension, for different values of \(\beta\). The trajectories are restricted to a square space of size \(2{,}000\) in the arbitrary units of the noise scale. In Fig.~\ref{fig:showcase} we present the distributions of horizontal positions from \(1{,}000\) of these trajectories. It is straightforward to see that white noise leads to an under-explored state space, since the trajectories remain too close to the center. On the other end of the spectrum, red noise is too correlated and thus the trajectories tend to get stuck in the edges. Intermediate values between \(\beta=0.5\) and \(\beta=1\) result in an adequate balance between these two effects. The developmentally inspired \(\beta\) values reported in Fig.~\ref{fig:behavior} fall within this range. However, white noise would be beneficial in a smaller space and red noise in a larger one. In sum, the optimal noise color will depend on the exploration requirements.

\subsection{Developmental Noise Schedule}

In Section \ref{sec:infant}, we showed that \(\beta\) increases with age, indicating a developmental shift toward increasingly temporally structured movements. Inspired by this finding, we introduce a developmental schedule that gradually increases the noise color during training for RL agents.

Concretely, for a noise block of length \(L\), we divide the total training time \(T\) into \(N=T/L\) discrete intervals of length \(L\). The value of \(\beta\) is kept constant during each of these intervals, allowing us to use the colored noise block mechanism explained before. However, \(\beta\) is scheduled throughout the entire training. Once a block's noise samples are exhausted at training time \(t\), a new one is generated with fixed color \(\beta\) given by:
\begin{equation} \label{eq:beta-schedule}
    \beta(t) = \beta_a + \beta_b \cdot \frac{t}{T},
\end{equation}

\noindent where the parameters \(\beta_a\) and \(\beta_b\) are estimated from the values of \(a\) and \(b\) obtained empirically (see Eq.~\ref{eq:beta-infant}). To estimate \(\beta_a\) and \(\beta_b\), we make a strong yet reasonable assumption. Recall that the infant data shows an increase from \(\beta\approx0.7\) at 8 weeks to \(\beta\approx0.9\) at 30 weeks. Extrapolating beyond this range would be too speculative. For example, there is extensive evidence showing that human adults display pink noise (\(\beta=1\)) across multiple behaviors~\cite{gilfriche2018frequency,hennig2011nature,duarte2001long}. Thus, we treat \(\beta=1\) as a likely upper bound on the developmental noise. For the sake of simplicity, we assume that the noise colors estimated from the video recordings are our best estimate of an infant's action noise throughout development, and thus we take this at face value.

In view of this, we define the \textit{baby noise} schedule as:
\begin{equation} \label{eq:beta-schedule}
    \beta(t) = 0.7 + 0.2\cdot \frac{t}{T}
\end{equation}

This baby noise function increases linearly with the fraction of training time $\nicefrac{t}{T}$ from \(0.7\) to \(0.9\) . At any time \(t\) during training, a noise block of length \(L\) can be generated using a fixed color \(\beta(t)\). 

\subsection{Integration into RL Algorithms}

Colored noise has been shown to benefit exploration in multiple RL algorithms, including DDPG~\cite{lillicrap2016continuous}, MPO~\cite{eberhard2023pink}, PPO~\cite{hollenstein2024colored}, and SAC~\cite{eberhard2023pink}. Here, we show how baby noise can be integrated into two different classes of algorithms.

For deterministic algorithms such as DDPG~\cite{lillicrap2016continuous} and TD3~\cite{fujimoto2018addressing}, exploration is performed by adding noise to the deterministic policy. More specifically, these algorithms explore by means of:
\begin{equation} \label{eq:deterministic-noise}
    a_t = \mu(s_t) + \sigma \odot \xi_t,
\end{equation}

\noindent where \(a_t\) is the selected action, \(\mu(s_t)\) is the deterministic policy at state \(s_t\), \(\sigma\) is a fixed standard deviation, and \(\xi_t \sim \mathcal{N}(0,I)\) is a Gaussian noise sample. We replace this white noise with the developmentally scheduled baby noise generated for the corresponding timestep.

On the other hand, SAC~\cite{haarnoja2018soft} is a stochastic algorithm that samples actions using the reparametrization trick:
\begin{equation} \label{eq:stochastic-noise}
    a_t = \tanh\left(\mu(s_t) + \sigma(s_t) \odot \xi_t\right),
\end{equation}

\noindent where \(\mu(s_t)\) and \(\sigma(s_t)\) are the mean and standard deviation of the policy and \(\tanh(\cdot)\) is used to generate a squashed Gaussian output. In line with~\cite{eberhard2023pink}, we replace the white noise \(\xi_t\) with our baby noise sampling while maintaining the original reparametrization.

The integration of our baby noise into these RL algorithms results in a curriculum over exploration dynamics, transitioning toward increasingly smoother and more structured exploratory behavior as training progresses.

\section{Infant-Inspired Action Noise in Deep RL} \label{sec:experiments}

We evaluate the proposed infant-inspired exploration strategy across several experiments designed to assess whether a scheduled temporally correlated action noise can improve exploration and learning compared to alternative baselines.

\subsection{Algorithms}

We integrate the baby noise into two widely used RL algorithms: TD3 and SAC. For TD3, exploration is implemented by adding noise directly to the deterministic policy output (Eq.~\ref{eq:deterministic-noise}). For SAC, the colored noise process is used in the policy distribution (Eq.~\ref{eq:stochastic-noise}). In both cases we replace the standard noise distributions from the algorithm implementations of the Stable-Baselines3 library~\cite{stable-baselines3}. The rest of the algorithm remains unchanged.

\subsection{Environments}

Experiments are conducted on a range of continuous-control environments. We use a total of 12 different environments from Gymnasium~\cite{towers2025gymnasium} and Gymnasium-robotics: four classic control environments, four locomotion environments, and four maze environments. These environments range from simple low-dimensional control tasks to high-dimensional problems, additionally requiring different degrees of exploration. See Fig.~\ref{fig:results} for details. 

\subsection{Noise types}

We compare our developmentally scheduled noise defined by Eq.~\ref{eq:beta-schedule} with six other exploration strategies: \(\beta=0\) (white noise), \(\beta=0.5\) (blush noise), $\beta~=~0.75$ (rose noise),  \(\beta=1\) (pink noise), \(\beta=2\) (red noise), and OU noise. White and colored noises are generated in blocks, as described in Section \ref{sec:action-noise}. OU noise is only used with the TD3 algorithm by means of its default implementation provided in Stable-Baselines3.

\subsection{Training}

For each environment, we perform a set of preliminary runs using white noise to estimate a reasonable training duration and set of algorithm parameters (eg. \(\sigma\) for TD3). The main interest of this study is in early exploration rather than final performance, so shorter training times are favored. All noise types are trained with the same hyperparameters, using the default values from Stable-Baselines3. Each triplet \{algorithm, environment, noise\} is trained using 10 different seeds.

\subsection{Evaluation}

We evaluate the training progress as the area under the learning curve (AUC) from the returns collected during each training episode. Since the rewards of different environments are scaled differently, we repeat the normalization protocol of~\cite{eberhard2023pink}. First, we compute the AUC for each run independently. For each pair \{algorithm, environment\}, we z-score the runs resulting from all noise types. The resulting normalized AUC can be aggregated and compared across algorithms and environments, allowing us to identify global performance differences between different noise types.

Additionally, since all the colored noises are generated through the same spectral shaping and we control the seeds used during training, we can pair runs from white and colored noises. This allows us to calculate a win rate for every colored noise, defined as the total fraction of runs with higher AUC than the paired white noise run. We perform significance tests of the win rates against chance level using a two-sided binomial test over all paired runs.

\subsection{Results}

\begin{figure}[!t]
  \centering
    \begin{subfigure}{0.255\textwidth}
        \centering\includegraphics[width=\textwidth]{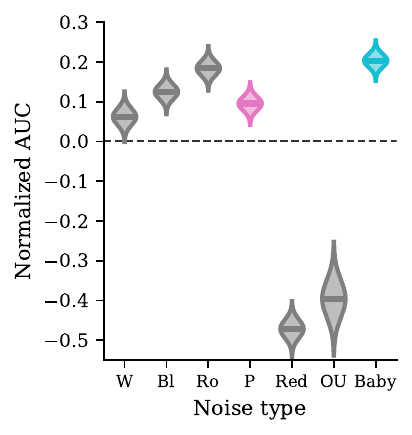}
        \caption{Normalized AUC scores across all algorithms and environments.}
    \end{subfigure}
    \hfill
    \begin{subfigure}{0.215\textwidth}
        \centering\includegraphics[width=\textwidth]{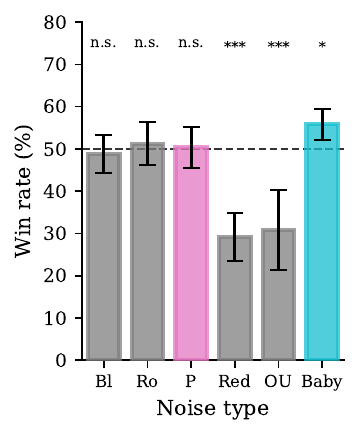}
        \caption{Win rates for colored noises vs. white noise.}
    \end{subfigure}
    \caption{Aggregated RL results. Violins and error bars indicate standard errors. Noise types compared are W:~white (\(\beta=0\)), Bl:~blush (\(\beta=0.5\)), Ro:~rose ($\beta=0.75$), P:~pink (\(\beta=1\)), Re:~red (\(\beta=2\)), OU and Baby (ours). Blush, rose, pink, and baby noises achieve higher normalized AUC than white noise, but only the latter significantly outperforms white noise over paired runs with a win rate of \(56\%\) (\(p<0.05\)).}
    \label{fig:results}
\end{figure}

The aggregated scores are shown in Fig.~\ref{fig:results}.A. Averaging over all algorithms and environments, the baby noise achieves the highest normalized AUC score, closely followed by the rose ($\beta=0.75$), blush ($\beta=0.5$) and pink ($\beta=1.0$) noises, all of which outperform white noise ($\beta=0$). The high temporal correlations induced by red and OU noise yield the worst performances. Additional details are provided in Table~\ref{tab:results}. Segregating the scores per algorithm reveals that baby noise achieves the highest performances for both SAC and TD3. The moderately colored blush, rose, and pink noises also outperform the white noise for SAC (as reported in \cite{eberhard2023pink}) but not for TD3.

A closer inspection of Table~\ref{tab:results} shows that our developmentally inspired exploration strategy achieves the best performance in only one of 24 conditions (\texttt{\small PointMaze\_UMaze-v3} trained with TD3). Its overall success stems from its consistency: only three scores below the zero-mean. By way of example, the baby noise is the best exploration strategy when averaging over all locomotion environments trained with SAC despite only being the second best for two out of four individual environments. Other temporally correlated noises dominate specific conditions, e.g., blush noise is consistently the best across maze environments with SAC. 

The consistency of the infant-inspired exploration is also reflected in the win rates shown in Fig.~\ref{fig:results}.B. This paired comparison provides a direct measure of the fraction of runs where each temporally correlated noise outperforms white noise. The baby noise is the only one with a win rate significantly above chance level \(56\%\) (\(p<0.05\)). On the other hand, the high AUC scores of the blush, rose, and pink noises do not translate into higher-than-chance win rates.

\section{Discussion and Future Work}

\begin{table*}[!t]
    \centering
\caption{Average AUC scores for different algorithms and environments, z-scored across multiple noises and seeds. Noise types compared are: white (\(\beta=0\)), blush (\(\beta=0.5\)), rose ($\beta=0.75$), pink (\(\beta=1\)), red (\(\beta=2\)), OU and Baby (ours). Standard errors are omitted due to limited space. The best result is highlighted in \textbf{bold} and the second best is \underline{underlined}.}
    \addtolength{\tabcolsep}{-0.1em}
    \begin{tabular}{lrrrrrrrrrrrrr}
    \toprule
     & \multicolumn{6}{c}{SAC} & \multicolumn{7}{c}{TD3} \\
     \cmidrule(lr){2-7}\cmidrule(lr){8-14}
    Environment & White & Blush & Rose & Pink & Red & {Baby} & White & Blush & Rose & Pink & Red & OU & Baby \\
    \midrule
    \texttt{MountainCarContinuous-v0} & --1.70 & --0.49 & 0.48 & \underline{0.69} & \textbf{0.79} & 0.23 & --0.39 & --0.39 & --0.39 & --0.40 & --0.44 & \textbf{2.39} & --\underline{0.39} \\
    \texttt{Pendulum-v1} & \underline{0.21} & 0.19 & 0.08 & \textbf{0.27} & --0.76 & 0.02 & \textbf{0.78} & \underline{0.51} & 0.39 & 0.37 & --0.34 & --2.18 & 0.46 \\
    \texttt{InvertedPendulum-v5} & \underline{0.39} & 0.06 & \textbf{0.59} & 0.27 & --1.43 & 0.13 & 0.21 & \textbf{0.62} & 0.49 & \underline{0.49} & --0.36 & --1.77 & 0.33 \\
    \texttt{InvertedDoublePendulum-v5} & 0.23 & \textbf{0.31} & 0.23 & --0.07 & --0.95 & \underline{0.25} & 0.02 & 0.18 & 0.19 & \textbf{0.80} & --0.43 & --1.29 & \underline{0.53} \\
    All classic control environments & --0.22& 0.02 & \textbf{0.34} & \underline{0.29} & --0.59 & 0.16 & 0.16 & 0.23 & 0.17 & \textbf{0.31} & --0.29 & --0.71 & \underline{0.23} \\
    \midrule
    \texttt{Hopper-v5} & --0.10 & --0.04 & 0.08 & \textbf{0.46} & --0.54 & \underline{0.14} & 0.18 & \textbf{0.55} & \underline{0.38} & --0.22 & 0.34 & --1.43 & 0.20 \\
    \texttt{Swimmer-v5} & --0.13 & \textbf{0.34} & 0.07 & --0.40 & --0.18 & 0.31 & \underline{0.31} & --0.18 & 0.19 & --0.16 & --0.98 & \textbf{0.43} & \underline{0.38} \\
    \texttt{HalfCheetah-v5} & \textbf{0.62} & --0.27 & --0.33 & --0.40 & 0.09 & \underline{0.29} & --0.27 & --0.20 & 0.34 & \textbf{0.49} & \underline{0.40} & --1.0 & 0.23 \\
    \texttt{Ant-v5} & --0.17 & --0.28 & \underline{0.07} & --0.45 & \textbf{0.86} & --0.03 & \textbf{0.76} & 0.07 & 0.31 & \underline{0.43} & 0.29 & --2.24 & 0.38 \\
    All locomotion environments & 0.05 & --0.06 & --0.03 & \underline{0.09} & 0.06 & \textbf{0.20} & 0.25 & 0.06 & \textbf{0.31} & 0.13 & 0.01 & --1.06 & \underline{0.30} \\
    \midrule
    \texttt{PointMaze\_Open-v3} &  --0.12 & \underline{0.36} & \textbf{0.61} & 0.46 & --1.56 & 0.25 & 0.30 & \textbf{0.58} & --0.31 & 0.05 & --0.95 & \underline{0.52} & --0.20 \\
    \texttt{PointMaze\_UMaze-v3} & 0.21 & \textbf{0.49} & 0.35 & 0.27 & --1.71 & \underline{0.39} & 0.26 & --0.37 & \underline{0.27} & --0.14 & --0.57 & 0.10 & \textbf{0.45} \\
    \texttt{PointMaze\_Medium-v3} & --0.08 & \textbf{0.49} & 0.23 & \underline{0.41} & --1.06 & 0.00 & 0.06 & 0.19 & 0.26 & --0.66 & --0.82 & \textbf{0.70} &  \underline{0.27} \\
    \texttt{PointMaze\_Large-v3} & --0.60 & \textbf{0.32} & \underline{0.28} & 0.16 & --0.23 & 0.07 & \underline{0.51} & --0.06 & --0.46 & --0.43 & --0.76 & \textbf{1.02} & 0.18 \\
    All maze environments & --0.15 & \textbf{0.42} & \underline{0.37} & 0.33 & --1.14 & 0.18 & \underline{0.28} & 0.09 & --0.06 & --0.29 & --0.77 & 0.18 & \textbf{0.59} \\
    \midrule
    All environments & --0.10 & 0.12 & \underline{0.19} & 0.18 & --0.56 & \textbf{0.21} & \underline{0.23} & 0.13 & 0.14 & 0.05 & --0.38 & --0.40 & \textbf{0.24}\\
    \bottomrule
    \end{tabular}
    \label{tab:results}
\end{table*}

% CONCLUSION
This work provides, to the best of our knowledge, the first quantitative characterization of the transition between stochastic and structured behaviors during early infancy making use of PSDs. Our results suggest that development involves not only changes in the nature of the behaviors but also in the structure of their variability. This development of a temporal auto-correlation may be indicative of a shift from exploratory spontaneous movements toward goal-directed behaviors~\cite{zettersten2025helpless} during the protracted helpless stage~\cite{cusack2024helpless}. 

The application of a developmentally scheduled exploration to deep RL algorithms yields higher performances than commonly used strategies, including Gaussian white noise and colored pink noise. Besides moderate temporal correlations, our \textit{baby noise} introduces a unique annealing of the noise color \(\beta\) that forces agents to continuously adapt to increasingly expanding state coverages. In light of our results, the next step is to confirm whether developmentally scheduled noise also improves exploration beyond the environments evaluated here. Of particular interest would be experiments involving infant embodiments, for example making use of MIMo's growing body \cite{mattern2024mimo,lopez2025mimo,philipp2026embodiment}.

This study can pave the way for more developmentally inspired learning strategies for deep RL. There is an ever-growing corpus of works that characterize the changes undergone by humans during the first few weeks and months of life, spanning from neurophysiology to behavior. Our results, as well as recent successes from infant-inspired representation models \cite{zaadnoordijk2022lessons,aubret2023time}, shine a promising light on the potential of developmentally grounded mechanisms for artificial agents. In turn, strengthening the connections between developmental science and artificial systems can lead to a deeper understanding of the developmental origins of the building blocks of intelligent behaviors.

%%%%%%%%%%%%%%%%%%%%%%%%%%%%%%%%%%%%%%%%%%%%%%%%%%%%%%%%%%%%%%%%%%%%%%%%%%%%%%%%


%%%%%%%%%%%%%%%%%%%%%%%%%%%%%%%%%%%%%%%%%%%%%%%%%%%%%%%%%%%%%%%%%%%%%%%%%%%%%%%%

\small 
\bibliographystyle{IEEEtran}
\bibliography{1_references}

\end{document}